\documentclass[sigconf]{acmart}

\usepackage{amsthm} 
\usepackage{subfigure}
\usepackage{natbib}
\usepackage{hyperref}
\usepackage{cleveref}
\usepackage[ruled]{algorithm2e}

\newtheorem{Assumption}{\bf Assumption}

\def \cL{\mathcal{L}}
\def \cO{\mathcal{O}}
 \def \cD{\mathcal{D}}
 
\AtBeginDocument{%
  }

\setcopyright{acmlicensed}
\copyrightyear{2025}
\acmYear{2025}
\setcopyright{acmlicensed}
\acmConference[KDD '25] {Proceedings of the 31st ACM SIGKDD Conference on Knowledge Discovery and Data Mining V.2}{August 3--7, 2025}{Toronto, ON, Canada.}
\acmBooktitle{Proceedings of the 31st ACM SIGKDD Conference on Knowledge Discovery and Data Mining V.2 (KDD '25), August 3--7, 2025, Toronto, ON, Canada}
\acmISBN{979-8-4007-1454-2/25/08}
\acmDOI{10.1145/3711896.3736832}

\begin{document}

%%
%% The "title" command has an optional parameter,
%% allowing the author to define a "short title" to be used in page headers.
\title{Addressing Correlated Latent Exogenous Variables in Debiased Recommender Systems}

%%
%% The "author" command and its associated commands are used to define
%% the authors and their affiliations.
%% Of note is the shared affiliation of the first two authors, and the
%% "authornote" and "authornotemark" commands
%% used to denote shared contribution to the research.
\author{Shuqiang Zhang}
\authornote{Shuqiang Zhang and Yuchao Zhang contributed equally to this research.}
\affiliation{%
  \institution{Dalhousie University}
  \city{Halifax}
  \country{Canada}
}
\email{sh673592@dal.ca}

\author{Yuchao Zhang}
\authornotemark[1]
\affiliation{%
  \institution{Beijing University of Chemical Technology}
  \city{Beijing}
  \country{China}
}
\email{2023030252@buct.edu.cn}

\author{Jinkun Chen}
\affiliation{%
  \institution{Dalhousie University}
  \city{Halifax}
  \country{Canada}
}
\email{jinkun.chen@dal.ca}

\author{Haochen Sui}
\authornote{Haochen Sui is the corresponding author.}
\affiliation{%
  \institution{University of Michigan}
  \city{Ann Arbor}
  \country{United States}
}
\email{hcsui@umich.edu}

%%
%% By default, the full list of authors will be used in the page
%% headers. Often, this list is too long, and will overlap
%% other information printed in the page headers. This command allows
%% the author to define a more concise list
%% of authors' names for this purpose.
\renewcommand{\shortauthors}{Shuqiang Zhang, Yuchao Zhang, Jinkun Chen, and Haochen Sui}

%%
%% The abstract is a short summary of the work to be presented in the
%% article.

% intro + related work 2.25, preliminary 0.75, ours, 2-2.5 experiment + conclusion 2.5, 3

\begin{abstract}
Recommendation systems (RS) aim to provide personalized content, but they face a challenge in unbiased learning due to selection bias, where users only interact with items they prefer. This bias leads to a distorted representation of user preferences, which hinders the accuracy and fairness of recommendations. To address the issue, various methods such as error imputation based, inverse propensity scoring, and doubly robust techniques have been developed. Despite the progress, from the structural causal model perspective, previous debiasing methods in RS assume the independence of the exogenous variables. In this paper, we release this assumption and propose a learning algorithm based on likelihood maximization to learn a prediction model. We first discuss the correlation and difference between unmeasured confounding and our scenario, then we propose a unified method that effectively handles latent exogenous variables. Specifically, our method models the data generation process with latent exogenous variables under mild normality assumptions. We then develop a Monte Carlo algorithm to numerically estimate the likelihood function. Extensive experiments on synthetic datasets and three real-world datasets demonstrate the effectiveness of our proposed method. The code is at \url{https://github.com/WallaceSUI/kdd25-background-variable}.

% Recommender systems are prone to being influenced by exogenous variables, such as selection bias introduced by observed exogenous confounders. While extensive research has been conducted to address the impact of these observed exogenous variables, there has been limited work on managing correlated latent exogenous variables. Existing approaches primarily focus on unmeasured confounding, utilizing methods like sensitivity analysis or relying on unbiased datasets to mitigate their effects. In this study, we propose a unified method that effectively handles latent exogenous variables, with unmeasured confoundingbeing a specific case of our approach. Our method models the data generation process—including latent exogenous variables—under mild assumptions, such as the additive and normality assumptions. We then develop a Monte Carlo algorithm to numerically estimate the likelihood function based on the symmetry of the endogenous variables. Extensive experiments on synthetic datasets and three real-world datasets demonstrate the effectiveness of our proposed method.
\end{abstract}

%%
%% The code below is generated by the tool at http://dl.acm.org/ccs.cfm.
%% Please copy and paste the code instead of the example below.
%%


%%
%% Keywords. The author(s) should pick words that accurately describe
%% the work being presented. Separate the keywords with commas.
%% A "teaser" image appears between the author and affiliation
%% information and the body of the document, and typically spans the
%% page.
% \begin{teaserfigure}
%   \includegraphics[width=\textwidth]{sampleteaser}
%   \caption{Seattle Mariners at Spring Training, 2010.}
%   \Description{Enjoying the baseball game from the third-base
%   seats. Ichiro Suzuki preparing to bat.}
%   \label{fig:teaser}
% \end{teaserfigure}

% \received{20 February 2007}
% \received[revised]{12 March 2009}
% \received[accepted]{5 June 2009}

%%
%% This command processes the author and affiliation and title
%% information and builds the first part of the formatted document.
\maketitle

\section{Introduction}

% 推荐系统是解决信息过载（information overload）的有效工具,在许多场景例如电商 [],social media [],娱乐 [] 等领域有很重要的应用。推荐系统旨在基于历史交互数据和用户物品特征预测用户对物品的偏好。However,一个重要的挑战是推荐系统收集到的训练数据是missing not at random (MNAR)的,是由于用户可以自我选择对哪些物品进行反馈导致的,例如用户可能更倾向于对他们感兴趣的物品评分,而对不感兴趣的物品不评分。(会导致训练数据的分布和target population的分布不一致,导致sub optimal的推荐表现。)带来的问题

% 为了解决这个问题,一类以往的研究提出使用Error imputation based方法,which first impute the missing ratings, then train a rating prediction model based on both observed rating and imputed rating。此外,还有一类方法使用倾向得分,即rating被观测到的概率,reweight observed rating来对齐观测数据和target population的分布。Doubly robust (DR) based方法combine imputation model和propensity model来训练预测模型,which is unbiased when either imputed errors or learned propensities are correct.

% 处理unmeasured confounding在推荐里是如何做的

% 外生变量的现实意义,外生变量的相关性的推荐系统现实场景（说明我们这个场景不是完全理论上考虑的,是现实生活中是真的有的）,和unmeasured confounding的区别和联系（shuqiang & yuchao & chunyuan）

% 写作的时候可以参考IPS,DR-JL,以及Addressing Unmeasured Confounder for Recommendation
% with Sensitivity Analysis来进行写作和逻辑整理

% The contribution of this paper can be summarized below

% 1:

Recommendation systems (RS) have become popular tools for providing personalized content across various domains, including education, e-commerce, and entertainment, as they effectively alleviate the information overload \cite{35,liu2024user,wang2024entire,li2025causal}. However, a significant issue arises from biases present in user interaction data, particularly selection bias \cite{32,33,34,zheng2025adaptive}. This bias stems from the fact that users tend to engage only with items that align with their preferences, leading to a distorted representation of their true interests. Such biases undermine the accuracy and fairness of recommendation models, posing a challenge to unbiased learning \cite{liu2022kdcrec,saito2019towards,39,li2023balancing,li2023removing}.

In response to these challenges, several strategies have been developed to mitigate biases and enhance the quality of recommendations. One prominent approach is the Error Imputation Based (EIB) method \cite{35,36}, which treats the selection bias as a data missing problem, filling in the gaps with pseudo labels for prediction model training. Furthermore, causality-based techniques~\citep{huang2025visual,huang2025text,zhou2025two,wang2025proximity} provide a promising direction for addressing these issues ~\citep{wang2023optimal, wang2024rcfr, chen2024improving, wang2025inter}. Inverse Propensity Scoring (IPS) \cite{37,38}, addresses selection bias by estimating the propensity score, which refers to the probability of a unit being observed and adjusting the weights of observed interactions to better reflect the true underlying distribution. Further advances have led to the development of the Doubly Robust (DR) method \cite{39,40,wang2024improving}, which combines both error imputation and inverse propensity weighting. This approach not only reduces bias but also enhances the robustness of the model, making it more reliable across a range of data scenarios. Building upon these foundational methods, recent research has proposed several improvements, including the integration of advanced propensity score learning strategy \cite{45,46}, the use of information-theoretic principles \cite{49,50}, and balancing of bias and variance \cite{41,42,wu2024invariant}.

% However, when modeling the treatment (propensity) and outcome using observed features, estimation errors may arise, with previous work attributing these errors to the presence of unmeasured confounding. For example, sensitivity analysis based methods are proposed to address this issue, which assumes the hidden confounding strength is bounded and attempts to account for worst-case scenarios to offer a robust framework. Another line of research focuses on leveraging external sources of information, such as Randomized Controlled Trials (RCT) data, to calibrate the propensity and imputation models for reducing the impact of unmeasured confounding.

\begin{figure}[t]
\centering
\includegraphics[width=0.45\textwidth]{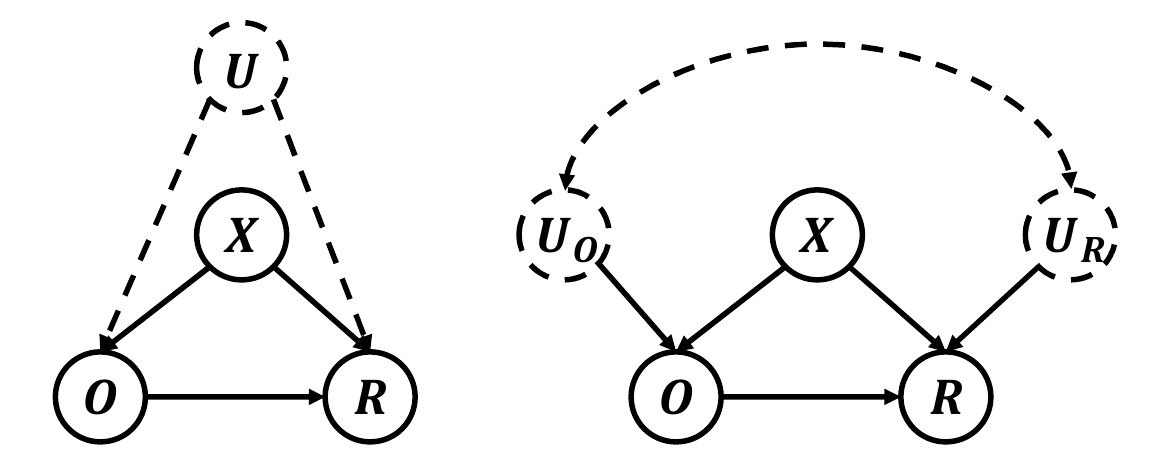}
\caption{Comparison between traditional unmeasured confounding (left) and latent exogenous variables (right). Dashed lines represent variables that are unobserved or latent.}
\label{fig_oag}
\vspace{-6pt}
\end{figure}

However, when modeling the treatment and outcome using observed features, previous methods assume the independence of exogenous variables. Specifically, the Structural Causal Model (SCM) of previous methods is defined by a triple $(U, V, F)$, where $V$ are observable endogenous variables (covariate $X$, the outcome observation indicator $O$, and outcome $R$) and $U$ are unobserved exogenous variables that cannot be caused by any variable in $V$, including $U_X, U_O ~\text{and}~U_R$. Previous methods assume that all the exogenous $U$ are required to be jointly independent. In this paper, we aim to relax the above independence assumption between exogenous variables, as shown in Figure \ref{fig_oag} (right).

Before introducing our method, we want to first discuss the correlation and difference of previous graphical models under traditional unmeasured confounding (Figure \ref{fig_oag} (left)) and scenario in our paper (Figure \ref{fig_oag} (right)). Specifically, for the traditional unmeasured confounding scenario, $U$ is an unmeasured \textbf{endogenous} variable, not an \textbf{exogenous} variable. In other words, the data generation process is $O=f_O(X, U, U_O)$ and $R=f_R(X, U, U_R)$ with unmeasured endogenous variable $U$, leading to a biased prediction if we only use variable $X$.
For dependent exogenous variables scenario, $U_O$ and $U_R$ are correlated exogenous variables of endogenous variables $O$ and $R$, i.e., $O=f_O(X, U_O)$ and $R=f_R(X, U_R)$ but with the correlated background variables $U_O$ and $U_R$.

In this paper, to train an unbiased prediction model in a correlated latent exogenous variables scenario, we propose a framework that incorporates exogenous variables into recommendation modeling, mitigating selection bias and improving robustness. Specifically, our method models the data generation process with latent exogenous variables under mild normality assumptions. We then develop a Monte Carlo algorithm to numerically estimate the likelihood function. Our contributions can be summarized below:
\begin{itemize}
    \item This paper considers a different unmeasured confounding case and discusses the correlation and difference with traditional unmeasured confounding scenario.
    % \vspace{3pt}
    \item We develop a Monte Carlo algorithm to numerically estimate the likelihood function based on the symmetry of the endogenous variables to handle latent exogenous variables.
    % \vspace{3pt} 
    \item Extensive experiments on both synthetic and real-world datasets demonstrate that our method successfully addresses the latent exogenous variable correlation problem.
\end{itemize}

% 3：Extensive experiments are conducted on real-world datasets, which demonstrates the effectiveness of our method.

\section{Related Work}

\subsection{Debiased Recommendation}

\subsubsection{Debiasing Methods without Unmeasured confounding}
Recommendation systems (RS) provide personalized recommendations for each user and have been widely applied in various fields such as education, commerce, and entertainment \cite{35}, often with time series data \cite{lai2024fts}, in order to address the information overload problem. However, the user interaction data inevitably contain biases, such as selection bias, where users interact only with items they prefer, resulting in a biased selection that challenges the unbiased learning \cite{32,33,34}. To address these issues, a line of methods are proposed to eliminate biases to ensure the fairness, accuracy, and reliability of the recommendation results, which are referred as debiased recommendation. Specifically, the Error Imputation Based (EIB) Estimators \cite{35,36} treat selection bias as a missing data problem and impute pseudo labels for prediction model learning. The Inverse Propensity Score (IPS) methods \cite{37,38,li2023propensity} aim to remove data biases by reweighting the observed events to restore the target distribution of all events. Combining both error imputation and inverse propensity reweighting, the Doubly Robust (DR) methods \cite{39,40,hxli2023litdr} effectively reduce the bias and achieve double robustness. Based on these fundamental methods, enhanced estimators are proposed, including combination with improved propensity model learning, integration with information-theoretic techniques \cite{49,50}, consideration the bias and variance trade-off \cite{41,42}, and leveraging a few unbiased Randomized Control Trial (RCT) data for model selection \cite{45,46, liu2022kdcrec}.

\subsubsection{Debiasing Methods with Unmeasured confounding}

While the aforementioned debiasing techniques significantly improve the accuracy and fairness of recommendation systems, they are often limited by the presence of unmeasured confounding variables—factors that simultaneously influence both the user’s interaction with items and the feedback they provide~\cite{ma2022learning,jiang2024confounder}. Confounding occurs when unobserved variables create spurious relationships between user preferences and the recommendation model, leading to biased outcomes even after applying debiasing methods. To address unmeasured confounding, various strategies have been proposed~\cite{1,xiao2024addressing}. One approach is to use sensitivity analysis, which attempts to account for worst-case scenarios by quantifying the potential impact of unmeasured confounding~\cite{1}. This method typically relies on optimization techniques that minimize the worst-case bias, offering a robust framework to cope with the hidden bias. However, the effectiveness of these methods hinges on assumptions about the nature of the hidden confounding, such as the belief that the true propensity scores can be bounded within the estimated values. When these assumptions are violated, the methods may fail to provide reliable, unbiased estimates. Another line of research focuses on leveraging external sources of information, such as controlled experimental data or Randomized Controlled Trials (RCTs), to calibrate the model and reduce the impact of unmeasured confounding~\cite{ma2022learning,xiao2024addressing}.

\subsection{Exogenous Variables}
Exogenous variables are factors that are independent of other system variables and are determined by external factors. They are typically treated as "inputs" or "noise sources" in causal models and are usually assumed to be independent random variables, with their values determined by external factors \cite{23,wang2025effective}. \citet{27} proposed the framework of causal models, highlighting the role of exogenous variables in causal inference. \citet{25} established that, under certain conditions, exogenous variables can be uniquely identified from observational data. \citet{26} further explored the identifiability issue, showing that nonlinear transformations can uniquely recover exogenous variables under specific conditions. \citet{29} introduced the theoretical framework of Normalizing Flows (NF), demonstrating its effectiveness in probabilistic modeling and inference, while \citet{24} propose causal Normalizing Flows framework combining NF with causal models, which can recover exogenous variables and perform causal inference. However, these transformations require strong model assumptions, and the correlation between exogenous variables can reduce the accuracy. In contrast, our method models the data generation process including exogenous variables to effectively address the correlation issue.

\section{Preliminary}
We first formulate debiasing problem: Let $\mathcal{U}=\{u_1, u_2, \dots, u_m\}$ denotes the set of $m$ users, $\mathcal{I}=\{i_1, i_2, \dots, i_n\}$ denotes the set of $n$ items, and $\cD = \mathcal{U} \times \mathcal{I}$ denotes the set of all user-item pairs. 
Denote $\mathbf{R}\in\{0,1\}^{m \times n}$ as the preference (e.g. rating or click) label matrix of all user-item pairs. Denote $x_{u,i}$ as the feature of user-item pair $(u,i)$. 
If $\mathbf{R}$ is fully observed, a  prediction model $\hat{r}_{u,i}=f(x_{u, i}; \theta)$ can be trained by minimizing the ideal loss
\begin{equation}
				\cL_{ideal}(\theta)  %=\cL_{ideal}(\phi)
				=  \frac{1}{|\cD|}  \sum_{(u,i) \in \cD} e_{u,i},  %= \bfE({e_{u,i}}),   
		\end{equation}    
where $e_{u,i}$ is the prediction error, such as the  cross entropy loss $e_{u,i} = CE(r_{u,i}, \hat r_{u,i}) =  -r_{u, i} \log \hat{r}_{u, i}-\left(1-r_{u, i}\right) \log \left(1-\hat{r}_{u, i}\right)$. 
% or squared loss 
% Take the least square loss as an example, 
% $e_{u,i} = (\hat r_{u,i}  - r_{u,i} )^2$.
In real recommendation applications,  users do not respond to all the items, and the response behavior is not purely random, resulting in the challenge of unbiased estimation of the ideal loss. Let $o_{u, i}=1$ if user $u$ response to item $i$, then $r_{u, i}$ with $o_{u, i}=0$ are not directly observable. Consequently, directly optimizing the ideal loss is not feasible. To solve this problem, a naive estimator optimizes the prediction model by minimizing the average prediction 
error corresponding to the observed preference
\begin{align}
\cL_{Naive}(\theta)  %=\cL_{ideal}(\phi)
				=  \frac{1}{|\cO|}  \sum_{(u,i) \in \cO} e_{u,i} =  \frac{1}{|\cO|}  \sum_{(u,i) \in \cD} o_{u, i}e_{u,i},
\end{align}
where $\cO = \{ (u,i) \mid (u,i)\in \cD, o_{u,i} =  1 \}$ is the set of observed user-item pairs. The Naive estimator is unbiased when the observed preference is missing at random (MAR). However, the presence of selection bias makes the data missing not at random (MNAR) and the observed preferences are no longer representative of all preferences. 

     % The alternative debiasing estimators
     The IPS method was proposed to re-weight the observed samples by $1/p_{u, i}$ to achieve unbiased ideal loss estimation, where the propensity $p_{u, i} = \operatorname{Pr}(o_{u,i}=1 | x_{u,i})$ denotes probability of a user $u$ response to an item $i$.
     % \footnote{\col{In this paper, \textit{propensity} and \textit{CTR} are interchangeable.}} 
     % in a setting of CVR prediction. 
     Then, the IPS estimator is given by
    \begin{align}
    \cL_{IPS}(\theta) = \frac{1}{|\cD|} \sum_{(u,i) \in \cD}  \frac{ o_{u,i}e_{u,i} }{ \hat p_{u, i} }, 
     \end{align}
where $\hat p_{u, i}=\pi(x_{u, i}; \psi)$ is the learned propensity. The IPS estimator is unbiased when the propensities of all user-item pairs are accurate, i.e., $\hat p_{u, i}=p_{u, i}$.  

The doubly robust (DR) estimator is given by 
         \begin{align}
         \cL_{DR}(\theta) = \frac{1}{|\cD|} \sum_{(u,i) \in \cD} \Big [ \hat e_{u,i}  +  \frac{ o_{u,i} (e_{u,i} -  \hat e_{u,i}) }{ \hat p_{u, i} } \Big ], 
     \end{align} 
which is an unbiased estimate of the ideal loss when either the imputed errors $\hat e_{u, i} = m(x_{u, i}; \phi)$ or the learned propensities $\hat p_{u, i}=\pi(x_{u, i}; \alpha)$ are correct, i.e., $\hat e_{u, i} = e_{u, i}$ or $\hat p_{u, i} = p_{u, i}$. DR method and its variants have demonstrated superior performance in debiasing recommendation.

When there is unmeasured confounding, several methods have been proposed to measure the effect of the unmeasured confounding. Ding et al. \citep{1} proposed to measure the strength of unmeasured confounding in the propensity using sensitivity analysis. They bound the uncertainty of the true propensity with a predefined degree $\Gamma$ with
\begin{equation}
    1+(1/\hat{p}_{u,i}-1)/\Gamma\le p_{u,i}\le 1+(1/\hat{p}_{u,i}-1)\Gamma,
\end{equation}
and train a prediction model by minimizing the corresponding loss under the worst propensity within the bound. Li et al. \citep{li2023balancing} proposed to leverage additional unbiased datasets without unmeasured confounding by minimizing 
\begin{equation}
    \mathcal{L}=\mathcal{L}_{Bias}+\mathcal{L}_{Unbias}+\mathcal{L}_(\hat{p},\tilde{p},\hat{\delta},\tilde{\delta}),
\end{equation}
where $\mathcal{L}_{Bias}$ is the IPS or DR loss trained on the biased dataset, $\mathcal{L}_{Unbias}$ is the IPS or DR loss trained on the unbiased dataset and $\mathcal{L}_(\hat{p},\tilde{p},\hat{\delta},\tilde{\delta})$ is some measure of distance between the propensities ($\hat{p},\tilde{p}$) trained on the two datasets, as well as the imputation ($\hat{\delta},\tilde{\delta}$) trained of the two datasets. However, Ding et al. consider a general case using sensitivity analysis, which cannot guarantee unbiasedness. In addition, Li et al. use additional RCT data to ensure unbiasedness, which is very expensive in real-world scenario. 
% \subsection{Methods without unmeasured confounding}
% Some previous approaches have addressed debiasing from the perspective of mitigating selection bias.
% \subsubsection{Inverse Propensity Score Estimator.}
% %逆倾向评分 (IPS)方法旨在通过对点击的事件进行加权来恢复所有事件的分布。我们
% % 第一个部分,without unmeasured confounding的方法,IPS,DR
% %问题定义,x_ui,
% % EIB, IPS, DR

% Firstly, we define Inverse Propensity Score

% \subsection{Methods with unmeasured confounding}
% 第二个部分,with unmeasured confounding的方, addressing

% 1: Sensivity analysis
% 2: data fusion, 融合有偏观测数据（多）和无偏数据（少）

% Lemma, previous methods for addressing unmeasured confounding are biased under our scenario
% \newpage

\section{Proposed Methods}
In this paper, we consider a more specific unmeasured confounding case, i.e., the bias caused by correlated exogenous variables in a parametric model, and propose a method that can achieve unbiasedness without additional RCT data. We will introduce the proposed method in this section. We begin by defining the data generation process. Then we derive the likelihood function when the preference is continuous. Finally, we extend our proposed method to the case where the preference is binary and we develop the Monte Carlo estimation algorithm.

\subsection{Problem Formulation}
Let $z_{u,i}$ represents the observational behavior of user-item pair $(u,i)$ such that $o_{u,i}=f_o(z_{u,i})$ where $f_o(z_{u,i})=1$ if $z_{u,i}>0$ and $f_o(z_{u,i})=0$ otherwise. Let $y_{u,i}$ represents the real preference of user-item pair $(u,i)$ such that $r_{u,i}=f_r(y_{u,i})$ where for continuous $r_{u,i}$, $f_r(y_{u,i})=y_{u,i}$ and for binary $r_{u,i}$, $f_r(y_{u,i})=1$ if $y_{u,i}>0$ and $f_r(y_{u,i})=0$ otherwise. As depicted in the right part of Figure \ref{fig_oag}, except from the observed exogenous variable $X$, we introduce two additional correlated latent exogenous variables $U_O$ and $U_R$ to measure the randomness in $O$ and $R$, respectively. For illustration purposes, we make the following normality assumption to model the correlation between $U_O$ and $U_R$, and the same modeling procedure can be applied to different distributions.
\begin{Assumption}
(Normality assumption) Let $U_O$ represents the unobserved exogenous variable that has an effect on $O$ and $U_R$ represent the unobserved exogenous variable that has an effect on $R$, then we assume
$U_O$ and $U_R$ follow the bi-variate normal distribution with mean 0, variance 1, and covariance $\rho$, i.e., $\begin{pmatrix}U_O\\U_R\end{pmatrix}\sim\mathcal{N}\left(\begin{bmatrix}
    0\\0
\end{bmatrix},\begin{bmatrix}1,\rho\\\rho,1\end{bmatrix}\right)$. \label{asm_n}\end{Assumption}
Then, let $g_o(x_{u,i};\theta_o)$ and $g_r(x_{u,i};\theta_r)$ be the functions for summarizing all the effect of $x_{u,i}$ on $o_{u,i}$ and $r_{u,i}$ respectively, i.e., 
\begin{align*}
        &z_{u,i}=g_o (x_{u,i};\theta_o)+U_{O,u,i},\\
        &y_{u,i}=g_r(x_{u,i};\theta_r)+U_{R,u,i},
\end{align*}where $\theta_o$ and $\theta_r$ are learnable parameters. With Assumption \ref{asm_n}, the joint distribution of $z_{u,i}$ and $y_{u,i}$, $f(z_{u,i},y_{u,i}|x_{u,i})$ is given by
% \begin{align}
%     &f(z_{u,i},y_{u,i}|x_{u,i})\notag\\
%     =&\frac{1}{2\pi\sqrt{1-\rho^2}}\exp\{-\frac{1}{2(1-\rho^2)}[(z_{u,i}-g_o(x_{u,i};\theta_o))^2\notag\\
%     &-2\rho(z_{u,i}-g_o(x_{u,i};\theta_o)(y_{u,i}-g_r(x_{u,i};\theta_r))\label{fn_pdf}\\
%     &+(y_{u,i}-g_r(x_{u,i};\theta_r))^2]\},\notag
% \end{align}
\begin{equation}
    f(z_{u,i},y_{u,i}|x_{u,i})=\frac{1}{2\pi|\Sigma|}\exp\{-\frac{1}{2}(\mathbf{w}_{u,i}-\boldsymbol{\mu}_{u,i})^T\Sigma^{-1}(\mathbf{w}_{u,i}-\boldsymbol{\mu}_{u,i})\},\label{fn_pdf}
\end{equation}
where $\mathbf{w}_{u,i}=\begin{bmatrix}
    z_{u,i}\\
    y_{u,i}
\end{bmatrix}$, $\boldsymbol{\mu}_{u,i}=\begin{bmatrix}
    g_o(x_{u,i};\theta_o)\\
    g_r(x_{u,i};\theta_r)
\end{bmatrix}$ and $\Sigma=\begin{bmatrix}
    1&\rho\\
    \rho&1
\end{bmatrix}$.
Then, the log-likelihood of the observed data is given by
\begin{align*}
    \mathcal{L}_{MLE}=&\sum_{\mathcal{O}}\log f(o_{u,i}=1,y_{u,i}|x_{u,i})+\sum_{\mathcal{D}\textbackslash\mathcal{O}}\log f(o_{u,i}=0|x_{u,i})\\
    =&\sum_{\mathcal{O}}\log f(z_{u,i}>0,y_{u,i}|x_{u,i})
    +\sum_{\mathcal{D}\textbackslash\mathcal{O}}\log f(z_{u,i}\le0|x_{u,i}).
\end{align*}
Then $\theta_o,\theta_r$ and $\rho$ can be obtained by maximizing $\mathcal{L}_{MLE}$, which requires an explicit formulation of the likelihood function. In the following two sections, we will discuss the scenarios where  $r_{u,i}$ is either continuous or binary.
\subsection{Continuous User Preference}
When $r_{u,i}$ describes some continuous metric of the user preference, such as watch time and purchase amount, the log-likelihood can be directly modeled as follows. First, since the log-likelihood of data in $\mathcal{D}\textbackslash\mathcal{O}$ only contains $z_{u,i}$ and by definition, the marginal distribution of $z_{u,i}$ is a standard normal distribution, it is easy to show that 
\begin{equation}
    f(z_{u,i}\le0|x_{u,i})=\Phi(-g_o(x_{u,i};\theta_o)),
\end{equation}
where $\Phi(\cdot)$ stands for the cumulative density function of standard normal distribution.
\par On the other hand, the log-likelihood of data in $\mathcal{O}$ can be obtained by integrate on $z_{u,i}$ over the joint pdf  described in \Cref{fn_pdf}. The result can be summarized as
\begin{align}
    &f(z_{u,i}>0,y_{u,i}=y|x_{u,i})\notag\\
    =&\frac{1}{\sqrt{2\pi}}\exp\{-\frac{(y_{u,i}-g_r(x_{u,i};\theta_r)^2)}{2}\}\label{fn_logl}\\
    &\cdot\Phi(\frac{g_o(x_{u,i};\theta_o)+\rho(y_{u,i}-g_r(x_{u,i};\theta_r))}{\sqrt{1-\rho^2}}),\notag
\end{align}
where a detailed derivation is given in \Cref{sec_ap_comp}. Thus, the full log-likelihood for continuous $r_{u,i}$ is given by
\begin{align*}
    \mathcal{L}_{MLE}=&\sum_{\mathcal{O}}-\frac{(y_{u,i}-g_r(x_{u,i};\theta_r)^2)}{2}\\
    &+\sum_{\mathcal{O}}\log\Phi(\frac{g_o(x_{u,i};\theta_o)+\rho(y_{u,i}-g_r(x_{u,i};\theta_r))}{\sqrt{1-\rho^2}})\\
    &+\sum_{\mathcal{D}\textbackslash\mathcal{O}}\log\Phi(-g_o(x_{u,i};\theta_o)),
\end{align*}
and the parameters $\theta_o,\theta_r$ and $\rho$ can be estimated by solving the following optimization problem,
\begin{equation}
    \theta_o,\theta_r,\rho=\arg\min_{\theta_o,\theta_r,\rho}-\mathcal{L}_{MLE}.
\end{equation}
\subsection{Monte Carlo Method for Estimating Likelihood Under Binary User Preference}
When we consider a more common scenario where $r_{u,i}$ represents a binary metric of user preference—such as click, purchase, tag, or comment behavior—the likelihood of the observed data is expressed as $f(z_{u,i}>0, y_{u,i}>0|x_{u,i})$ and $f(z_{u,i}>0, y_{u,i}\le0|x_{u,i})$. Under this circumstance, computing the log-likelihood becomes tedious, as it involves integrating over both $z_{u,i}$ and $y_{u,i}$ using the joint probability density function described in \Cref{fn_pdf}, which lacks a closed-form solution.
\par 
To estimate the likelihood function under binary user preference, we propose to use Monte Carlo method to numerically estimate the likelihood function. We begin with the likelihood function in \Cref{fn_logl} for continuous user preference and integrate $z_{u,i}$ to get $f(z_{u,i}>0, y_{u,i}>0|x_{u,i})$. The integration result for data with $r_{u,i}=1$ is given by
\begin{align}
&f(z_{u,i}>0, y_{u,i}>0|x_{u,i})\notag\\
    =&\mathbb{E}_{\epsilon}\left[\Phi(\frac{g_r(x_{u,i};\theta_r)+\rho \epsilon}{\sqrt{1-\rho^2}})I\{\epsilon>-g_o(x_{u,i};\theta_o)\}\right],
\end{align}
where $\epsilon\sim N(0,1)$ and $I(x)$ is the indicator function of $x$ such that $I(x)=1$ if $x$ is true and $I(x)=0$ otherwise. A detailed derivation is given in \Cref{sec_ap_comp}. To estimate this expectation, we propose to use Monte Carlo method. Specifically, we sample $\epsilon_l\sim N(0,1)$ repeatedly for $L$ times and approximate the expectation by
\begin{equation}
    MC^r_{u,i}=\frac{1}{L}\sum_{l=1}^L\left[\Phi(\frac{g_r(x_{u,i};\theta_r)+\rho \epsilon_l}{\sqrt{1-\rho^2}})I\{\epsilon_l>-g_o(x_{u,i};\theta_o)\}\right].\label{fn_mc}
\end{equation}
Then for $r_{u,i}=0$, by the rule of total probability, we have $f(z_{u,i}>0, y_{u,i}\le0|x_{u,i})=f(z_{u,i}>0|x_{u,i})-f(z_{u,i}>0, y_{u,i}>0|x_{u,i})$ and $f(z_{u,i}>0|x_{u,i})$ can be easily estimated using its marginal distribution as $f(z_{u,i}>0|x_{u,i})=\Phi(g_o(x_{u,i};\theta_o))$. Then, the log-likelihood function in this case can be summarized as
\begin{align*}
    \mathcal{L}_{MLE}^B=&\sum_{\mathcal{R}}\log MC^r_{u,i}
    +\sum_{\mathcal{O}\textbackslash\mathcal{R}}\log\left\{\Phi(g_o(x_{u,i};\theta_o))-MC^r_{u,i}\right\}\\
    &+\sum_{\mathcal{D}\textbackslash\mathcal{O}}\log\Phi(-g_o(x_{u,i};\theta_o)),
\end{align*}
where $\mathcal{R}=\{(u,i)|o_{u,i}=1, r_{u,i}=1\}$.
\subsection{Symmetry Property and Joint Learning of the Likelihood Function}
One particular problem associated with the optimization of $\mathcal{L}_{MLE}^B$ is the existence of the indicator function $I(\cdot)$, which is not differentiable when performing back propagation. One possible solution is to use some differentiable functions to replace $I(x)$. However, the problem is that only samples close to the cutting point can be used in the calculation of gradient, which may cause gradient explosion for samples close to the cutting point and gradient diminish for samples far away from the cutting point. Thus, we propose to leverage the symmetry property of the likelihood function and to optimize the likelihood function sequentially.
\par We first observe that the formulation of $f(z_{u,i}>0, y_{u,i}>0|x_{u,i})$ is symmetric, which means that if we change the order of integration, the result will have the same structure as \Cref{fn_mc} with $g_r$ and $g_o$ interchanged. Thus, the resulting Monte Carlo estimates will be
\begin{equation}
    MC^o_{u,i}=\frac{1}{L}\sum_{l=1}^L\left[\Phi(\frac{g_o(x_{u,i};\theta_o)+\rho \epsilon_l}{\sqrt{1-\rho^2}})I\{\epsilon_l>-g_r(x_{u,i};\theta_r)\}\right].\label{fn_mcd}
\end{equation}
where $g_o$ is out of the indicator function. Thus, we can first optimize $\theta_r$ using \Cref{fn_mc} with $\theta_o$ fixed and then optimize $\theta_o$ using \Cref{fn_mcd} with $\theta_r$ fixed. Thus, we decompose $\mathcal{L}_{MLE}^B$ into two parts as 
\begin{equation*}
    \mathcal{L}_{MLE}^R=\sum_{\mathcal{R}}\log MC^r_{u,i}
    +\sum_{\mathcal{O}\textbackslash\mathcal{R}}\log\left\{\Phi(g_o(x_{u,i};\theta_o))-MC^r_{u,i}\right\},
\end{equation*}
and the part corresponds to the observation
\begin{align*}
    \mathcal{L}_{MLE}^D=&\sum_{\mathcal{R}}\log MC^o_{u,i}
    +\sum_{\mathcal{O}\textbackslash\mathcal{R}}\log\left\{\Phi(g_o(x_{u,i};\theta_o))-MC^o_{u,i}\right\}\\
    &+\sum_{\mathcal{D}\textbackslash\mathcal{O}}\log\Phi(-g_o(x_{u,i};\theta_o)),
\end{align*}
and the  optimization problem can be formulated as optimizing the following two goals iteratively,
\begin{equation}
    \min_{\theta_o,\rho} -\mathcal{L}_{MLE}^D\text{ and }
     \min_{\theta_r,\rho} -\mathcal{L}_{MLE}^R.
\end{equation}
\subsection{Link to Loss Based Methods}
So far, we have derived the likelihood based optimization method, which accounts for the correlated latent exogenous variables for modeling user preference in recommender systems. On the other hand, there are abundant debiasing methods with corresponding loss functions. Thus, to leverage the advantages of existing works, we propose to include another loss term in the optimization objective to form a multi-task loss $\mathcal{L}_E$. Let $\mathcal{L}_{Debias}$ be the loss function of an arbitrary debiasing method, then the optimization objective is given by
\begin{equation}
    \mathcal{L}_E=\alpha(-\mathcal{L}_{MLE}^B)+(1-\alpha)\mathcal{L}_{Debias},\label{fn_alpha}
\end{equation}
where $\alpha\in(0,1)$ is a hyper-parameter that balances the likelihood and the debiasing loss. We give the pseudo-code of our proposed method in Algorithm \ref{alg2}.
\begin{algorithm}[t]
\caption{Proposed method for debiasing latent exogenous variables.}
\label{alg2}
\LinesNumbered
\KwIn{Propensity model $g_o(\theta_o)$; Prediction model $g_r({\theta_r})$;  Datasets $\mathcal{D}$; hyper-parameter $\alpha$. }
%\KwOut{output result}
Randomly initialize $\theta_o,\theta_r,\rho$\;
\While{not convergent}{
\For{number of steps for training the prediction model}{
Sample a batch of user-item pairs $(u,i)$ from $\mathcal{O}$;
Optimize $\theta_r$ and $\rho$ by minimizing $\alpha(-\mathcal{L}_{MLE}^R)+(1-\alpha)\mathcal{L}_{Debias}$;
}
\For{number of steps for training the propensity model}{
Sample a batch of user-item pairs $(u,i)$ from $\mathcal{D}$;
Optimize $\theta_o$ and $\rho$ by minimizing $-\mathcal{L}_{MLE}^D$;
}

$Epoch\leftarrow Epoch+1$\;
}
\KwOut{$g_o(\theta_o),g_r({\theta_r})$.}
\end{algorithm}
\begin{table*}[htbp]
  \centering
  \caption{Performance on MSE on the synthetic dataset with different values of $\rho$}
  \vspace{-8pt}
        \resizebox{0.9\linewidth}{!}{
    \begin{tabular}{l|ccccccccc}
    \toprule
          &  \multicolumn{9}{c}{\textsc{$\rho$}} \\ \midrule
    Method & -0.8   & -0.6 & -0.4 & -0.2   & 0 & 0.2 & 0.4   & 0.6 & 0.8 \\ \midrule

            NCF & 7.592$_{\pm2.536}$ & 4.785$_{\pm1.284}$ & 3.379$_{\pm1.525}$ & 2.033$_{\pm0.658}$ & 1.727$_{\pm0.590}$ & 1.707$_{\pm0.231}$ & 2.430$_{\pm0.544}$ & 3.614$_{\pm1.247}$ & 5.449$_{\pm1.594}$ \\
            
            EIB & 7.437$_{\pm1.652}$ & 4.659$_{\pm1.095}$ & 3.415$_{\pm0.963}$ & 2.337$_{\pm0.973}$ & 1.666$_{\pm0.429}$ & 1.636$_{\pm0.174}$ & 2.175$_{\pm0.440}$ & 3.554$_{\pm1.255}$ & 5.505$_{\pm1.719}$ \\
            
            IPS & 8.471$_{\pm1.205}$ & 5.595$_{\pm1.039}$ & 3.414$_{\pm0.530}$ & 2.566$_{\pm0.677}$ & 1.550$_{\pm0.183}$ & 1.567$_{\pm0.098}$ & 2.108$_{\pm0.124}$ & 3.406$_{\pm0.251}$ & 5.393$_{\pm1.555}$ \\
            
            DR & 5.518$_{\pm1.051}$ & 3.685$_{\pm0.570}$ & 2.491$_{\pm0.440}$ & 1.779$_{\pm0.258}$ & 1.416$_{\pm0.074}$ & 1.561$_{\pm0.189}$ & 2.452$_{\pm0.671}$ & 3.258$_{\pm0.664}$ & 5.287$_{\pm1.292}$ \\\midrule

            Ours & 1.481$_{\pm0.135}$ & 1.743$_{\pm0.593}$ & 2.149$_{\pm0.715}$ & 1.998$_{\pm0.364}$ & 1.662$_{\pm0.156}$ & 1.586$_{\pm0.119}$ & 1.625$_{\pm0.297}$ & 1.516$_{\pm0.112}$ & 1.459$_{\pm0.115}$ \\

        \bottomrule
    \end{tabular}}%
  \label{tab_mse}%
\end{table*}%
\section{Semi-Synthetic Experiments}

\subsection{Dataset and Pre-Processing}
%  MovieLens-100k
We conduct semi-synthetic experiments using the MovieLens 100K\footnote{https://grouplens.org/datasets/movielens/100k/} (\textbf{ML-100K}) dataset, focusing on the following two research questions (RQs): \textbf{RQ1}. Does the proposed method result in more accurate
estimation of user preference compared to the previous method in the presence of correlated latent exogenous variables when the user preference is continuous? \textbf{RQ2}. How accurate can we estimate the true correlation between the correlated latent exogenous variables?
\subsubsection{Experimental Setup} The \textbf{ML-100K} dataset contains 100,000 missing-not-at-random (MNAR) ratings from 943 users to 1,682 movies. Similar to the previous studies \cite{38,39}, we first compute the effect of the observed variable $\textbf{X}$ by fitting a Matrix Factorization (MF) model \cite{koren2009matrix}. Then, for each user-item pair $(u,i)$ and a given correlation $\rho\in\{-0.8,-0.6,-0.4,-0.2,0,0.2,0.4,0.6,0.8\}$, we generate two standard normal distributed error terms $\epsilon_{u,i}$ and $\delta_{u,i}$ with $cov(\epsilon_{u,i},\delta_{u,i})=\rho$. Then, to generate the true user preference $r_{u,i}$, we set $r_{u,i}=5x_{u,i}+\delta_{u,i}$. To generate the observation status, we first set $z_{u,i}=5*\tanh(x_{u,i}-\bar{x})+\epsilon_{u,i}-\beta$, where $\beta$ is a threshold parameter that controls the sparse rate of the semi-synthetic dataset. We set $o_{u,i}=1$ if $z_{u,i}>0$ according to the data generation process. In our experiment, we set $\beta=3$ to maintain a data sparsity rate of approximately 0.05. This results in a training set containing 88,010 samples. We select the size of the test set to be half of the training set, randomly choosing 44,005 samples from the entire set of user-item pairs to form the unbiased test set.

\subsection{Model Prediction Performance Under Different Values of $\rho$ (RQ1)}
To evaluate the predictive performance of our proposed method and compare its results with existing debiasing methods, we conducted experiments using different values of $\rho$ and summarized the results in \Cref{tab_mse}. We used NCF \cite{he2017neural}, EIB \cite{36}, IPS \cite{38}, and DR \cite{39} as baseline methods. From \Cref{tab_mse}, we can draw the following conclusions. First, all baseline methods perform well when $\rho$ is small, but their performance drops dramatically as $\rho$ increases. This finding highlights the ineffectiveness of conventional debiasing methods in the presence of correlated latent exogenous variables and the need for developing new methods that can handle the correlated latent exogenous variables. Second, our proposed method demonstrates stable performance across different $\rho$ values and significantly outperforms the baseline methods when $\rho$ is large. This indicates our method's capability to effectively address the challenge posed by correlated latent exogenous variables compared to existing methods. Third, there is a tendency for the mean squared error (MSE) to increase as $\rho$ becomes smaller. During these conditions, the IPS and DR methods outperform our proposed method. This is because that when $\rho$ is small, the bias introduced by the correlated latent exogenous variables is also small that can not dominate the total bias. This finding supports the incorporation of conventional debiasing methods to enhance predictive performance by balancing between our likelihood approach and the debiasing loss under varying $\rho$.
\subsection{Estimation of $\rho$ (RQ2)}
\begin{figure}[t]
\centering
\includegraphics[width=0.45\textwidth]{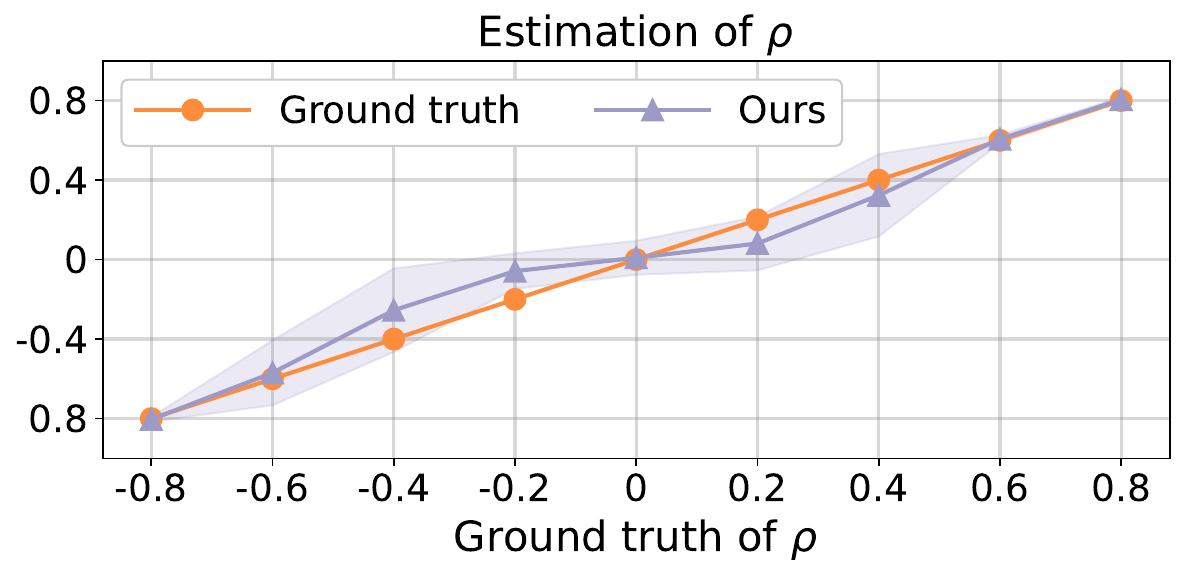}
\vspace{-6pt}
\caption{Estimation of $\rho$ with different ground truth.}
\label{fig_rho}
\vspace{-4pt}
\end{figure}

To evaluate whether our proposed method can recover the true value of $\rho$ from the observed data, we summarized the estimated value of $\rho$ in \Cref{fig_rho} with different ground truth values. We can draw the following two conclusions. First, our methods can recover the value of $\rho$ accurately, especially when the absolute true value of $\rho$ is large. Second, when $|\rho|$ is small, the variance of the estimated $\rho$ is larger than small value of $|\rho|$. This may suggest that when the magnitude of $\rho$ is small, the correlation information contained in the observed data is less, which may damage model performance and highlights the incorporation of conventional debiasing method to benefit from its debiasing ability.
% \subsection{}

\section{Real-World Experiments}
\begin{table*}[t]
  \centering
  \caption{Performance on AUC, Recall@K, and NDCG@K on the unbiased test set of Coat, Yahoo! R3, and KuaiRec. The best results are bolded, and the best baseline is underlined. * means statistical significant under p < 0.05 using pairwise t-test.}
  \vspace{-6pt}
    \resizebox{0.99\linewidth}{!}{
    \begin{tabular}{l|ccc|ccc|ccc}
    \toprule
          & \multicolumn{3}{c|}{\textsc{Coat}} & \multicolumn{3}{c|}{\textsc{Yahoo! R3}} & \multicolumn{3}{c}{\textsc{KuaiRec}} \\ \midrule
    Method & AUC   & R@5 & N@5 & AUC   & R@5 & N@5 & AUC   & R@50 & N@50 \\ \midrule

        NCF & 0.762$_{\pm0.011}$ & 0.441$_{\pm0.010}$ & 0.623$_{\pm0.011}$ & 0.682$_{\pm0.001}$ & 0.451$_{\pm0.002}$ & 0.674$_{\pm0.002}$ & 0.835$_{\pm0.001}$ & 0.691$_{\pm0.007}$ & 0.643$_{\pm0.004}$ \\

        CVIB & 0.759$_{\pm0.002}$ & 0.455$_{\pm0.006}$ & 0.636$_{\pm0.003}$ & 0.696$_{\pm0.001}$ & 0.452$_{\pm0.003}$ & 0.683$_{\pm0.002}$ & 0.792$_{\pm0.003}$ & 0.641$_{\pm0.004}$ & 0.584$_{\pm0.004}$ \\

        DIB & 0.747$_{\pm0.004}$ & 0.455$_{\pm0.007}$ & 0.643$_{\pm0.004}$ & 0.697$_{\pm0.003}$ & 0.436$_{\pm0.006}$ & 0.671$_{\pm0.005}$ & 0.800$_{\pm0.006}$ & 0.643$_{\pm0.006}$ & 0.576$_{\pm0.005}$ \\ \midrule

        IPS & 0.758$_{\pm0.006}$ & 0.448$_{\pm0.012}$ & 0.636$_{\pm0.011}$ & 0.690$_{\pm0.004}$ & 0.452$_{\pm0.004}$ & 0.674$_{\pm0.004}$ & 0.836$_{\pm0.004}$ & 0.692$_{\pm0.010}$ & 0.647$_{\pm0.005}$ \\

        SNIPS & 0.759$_{\pm0.011}$ & 0.449$_{\pm0.006}$ & 0.644$_{\pm0.008}$ & 0.692$_{\pm0.001}$ & 0.450$_{\pm0.002}$ & 0.677$_{\pm0.002}$ & 0.834$_{\pm0.001}$ & {0.703$_{\pm0.001}$} & 0.650$_{\pm0.001}$ \\

        AS-IPS & 0.759$_{\pm0.005}$ & 0.446$_{\pm0.004}$ & 0.633$_{\pm0.006}$ & 0.688$_{\pm0.007}$ & 0.454$_{\pm0.009}$ & 0.674$_{\pm0.007}$ & 0.836$_{\pm0.004}$ & 0.694$_{\pm0.004}$ & 0.644$_{\pm0.003}$ \\

        IPS-V2 & 0.750$_{\pm0.007}$ & 0.441$_{\pm0.004}$ & 0.619$_{\pm0.005}$ & 0.694$_{\pm0.007}$ & 0.456$_{\pm0.003}$ & 0.673$_{\pm0.005}$ & 0.838$_{\pm0.005}$ & 0.698$_{\pm0.006}$ & 0.649$_{\pm0.005}$ \\

        Multi-IPS & 0.765$_{\pm0.009}$ & 0.451$_{\pm0.014}$ & 0.635$_{\pm0.013}$ & 0.693$_{\pm0.004}$ & 0.453$_{\pm0.006}$ & 0.673$_{\pm0.006}$ & 0.837$_{\pm0.005}$ & 0.700$_{\pm0.002}$ & 0.650$_{\pm0.007}$ \\

        ESCM2-IPS & 0.763$_{\pm0.006}$ & 0.464$_{\pm0.010}$ & 0.647$_{\pm0.012}$ & 0.695$_{\pm0.007}$ & 0.459$_{\pm0.008}$ & 0.686$_{\pm0.009}$ & 0.841$_{\pm0.001}$ & 0.698$_{\pm0.004}$ & 0.646$_{\pm0.002}$ \\ \midrule

        DR-JL & 0.768$_{\pm0.011}$ & 0.446$_{\pm0.005}$ & 0.633$_{\pm0.007}$ & 0.696$_{\pm0.003}$ & 0.453$_{\pm0.003}$ & 0.678$_{\pm0.002}$ & 0.837$_{\pm0.001}$ & 0.698$_{\pm0.003}$ & 0.649$_{\pm0.003}$ \\

        MRDR-DL & 0.768$_{\pm0.009}$ & 0.457$_{\pm0.012}$ & 0.642$_{\pm0.013}$ & 0.695$_{\pm0.002}$ & 0.453$_{\pm0.003}$ & 0.679$_{\pm0.003}$ & 0.835$_{\pm0.002}$ & 0.697$_{\pm0.005}$ & 0.649$_{\pm0.003}$ \\

        DR-BIAS & 0.763$_{\pm0.010}$ & 0.449$_{\pm0.005}$ & 0.639$_{\pm0.007}$ & 0.699$_{\pm0.001}$ & 0.456$_{\pm0.003}$ & 0.677$_{\pm0.003}$ & 0.833$_{\pm0.004}$ & 0.687$_{\pm0.007}$ & 0.641$_{\pm0.002}$ \\

        DR-MSE & 0.764$_{\pm0.011}$ & 0.452$_{\pm0.005}$ & 0.635$_{\pm0.007}$ & 0.698$_{\pm0.003}$ & 0.456$_{\pm0.002}$ & 0.684$_{\pm0.002}$ & 0.833$_{\pm0.003}$ & 0.692$_{\pm0.005}$ & 0.643$_{\pm0.005}$ \\

        TDR-JL & \underline{0.770$_{\pm0.008}$} & 0.455$_{\pm0.017}$ & {0.653$_{\pm0.024}$} & {0.701$_{\pm0.004}$} & 0.456$_{\pm0.005}$ & 0.684$_{\pm0.006}$ & 0.839$_{\pm0.002}$ & 0.696$_{\pm0.004}$ & {0.651$_{\pm0.009}$} \\

        DR-V2 & 0.761$_{\pm0.004}$ & 0.452$_{\pm0.007}$ & 0.635$_{\pm0.005}$ & 0.690$_{\pm0.004}$ & 0.451$_{\pm0.004}$ & 0.682$_{\pm0.005}$ & 0.836$_{\pm0.003}$ & 0.700$_{\pm0.007}$ & 0.649$_{\pm0.005}$ \\

        Multi-DR & 0.758$_{\pm0.014}$ & 0.438$_{\pm0.016}$ & 0.627$_{\pm0.018}$ & 0.685$_{\pm0.004}$ & 0.453$_{\pm0.005}$ & {0.690$_{\pm0.005}$} & \underline{0.842$_{\pm0.002}$} & 0.700$_{\pm0.002}$ & {0.651$_{\pm0.005}$} \\

        RD-DR & 0.767$_{\pm0.005}$ & 0.465$_{\pm0.006}$ & 0.662$_{\pm0.005}$ & \underline{0.704$_{\pm0.006}$} & 0.460$_{\pm0.005}$ & 0.688$_{\pm0.004}$ & 0.837$_{\pm0.003}$ & 0.696$_{\pm0.005}$ & 0.645$_{\pm0.004}$ \\
        
        BRD-DR & 0.762$_{\pm0.004}$ & \underline{0.474$_{\pm0.005}$} & \underline{0.672$_{\pm0.005}$} & 0.703$_{\pm0.005}$ & 0.462$_{\pm0.004}$ & 0.691$_{\pm0.003}$ & 0.838$_{\pm0.004}$ & 0.704$_{\pm0.006}$ & 0.647$_{\pm0.004}$ \\
        
        ESCM2-DR & \underline{0.770$_{\pm0.006}$} & 0.460$_{\pm0.006}$ & {0.653$_{\pm0.009}$} & 0.700$_{\pm0.006}$ & {0.464$_{\pm0.005}$} & {0.690$_{\pm0.006}$} & 0.841$_{\pm0.004}$ & 0.700$_{\pm0.002}$ & 0.647$_{\pm0.008}$ \\

        KBDR & 0.769$_{\pm0.008}$ & 0.453$_{\pm0.005}$ & 0.640$_{\pm0.006}$ & 0.692$_{\pm0.007}$ & 0.453$_{\pm0.008}$ & 0.676$_{\pm0.004}$ & 0.838$_{\pm0.006}$ & 0.699$_{\pm0.008}$ & {0.651$_{\pm0.004}$} \\
        
        DCE-DR & \underline{0.770$_{\pm0.007}$} & 0.466$_{\pm0.003}$ & 0.662$_{\pm0.004}$ & 0.699$_{\pm0.009}$ & 0.459$_{\pm0.004}$ & 0.686$_{\pm0.003}$ & 0.828$_{\pm0.009}$ & 0.693$_{\pm0.005}$ & {0.644$_{\pm0.003}$} \\
        
         D-DR & 0.766$_{\pm0.007}$ & 0.466$_{\pm0.006}$ & 0.651$_{\pm0.006}$ & 0.699$_{\pm0.004}$ & \underline{0.465$_{\pm0.007}$} & \underline{0.692$_{\pm0.002}$} & 0.838$_{\pm0.004}$ & \underline{0.706$_{\pm0.006}$} & \underline{0.654$_{\pm0.003 }$} \\

        \midrule
        
        Ours-Naive & \textbf{0.775$_{\pm0.004}$} & {0.465$_{\pm0.009}$} & {0.658$_{\pm0.012}$} & {0.702$_{\pm0.005}$} & {0.464}$_{\pm0.001}$ & {0.691$_{\pm0.002}$} & \textbf{0.854$^{*}$$_{\pm0.003}$} & \textbf{0.721$^{*}$$_{\pm0.007}$} & {0.664$_{\pm0.004}$} \\

        % Ours-IPS & {0.757$_{\pm0.005}$} & {0.462$_{\pm0.012}$} & {0.638$_{\pm0.014}$} & {0.698$_{\pm0.003}$} & {0.452}$_{\pm0.003}$ & {0.675$_{\pm0.003}$} & {0.843$_{\pm0.003}$} & {0.727$_{\pm0.007}$} & {0.655$_{\pm0.004}$} \\

        Ours-DR & \textbf{0.775$_{\pm0.005}$} & \textbf{0.485$^{*}$$_{\pm0.012}$} & \textbf{0.679$^{*}$$_{\pm0.014}$} & \textbf{0.715$^{*}$$_{\pm0.003}$} & \textbf{0.486}$^{*}$$_{\pm0.003}$ & \textbf{0.713$^{*}$$_{\pm0.003}$} & {0.853$_{\pm0.004}$} & {0.715$_{\pm0.004}$} & \textbf{0.666$^{*}$$_{\pm0.006}$} \\

        \bottomrule
    \end{tabular}}%
  \label{tab_main}%
\end{table*}%
\subsection{Dataset}
\textbf{Coat} \cite{38}. This dataset contains MNAR data in the context of users purchasing coats from an online store. A total of 290 consumers were asked to rate 24 self-selected coats and 16 randomly chosen coats on a 5-point scale, in 300 items, resulting in 6,960 MNAR ratings in training set and 4,640 MAR ratings in test set. We binarize the ratings less than three to 0, otherwise to 1.\\
\textbf{Yahoo! R3} \cite{35}. The training set of this dataset consists of ratings from 15,400 users, with over 300,000 MNAR rating entries. The test set is collected by asking 5,400 users to rate 10 randomly displayed music, making it MAR data. We also binarize this 5-point scale dataset using the same criteria like \textbf{Coat} dataset.\\
\textbf{KuaiRec} \cite{gao2022kuairec}. This dataset is an industrial dataset with 4,676,570 video watching ratio from 1,411 users to 3,327 items. For this dataset, we binarize the ratios less than two to 1, otherwise to 0.

% Coat, Yahoo! R3
\subsection{Experiment Protocol}
\subsubsection{Baselines.}  We consider the following debiasing baselines, including: (1) No propensity: \textbf{NCF} \cite{he2017neural}, \textbf{CVIB} \cite{49}, and \textbf{DIB} \cite{2}; (2) IPS-based: \textbf{IPS}~\cite{38}  \textbf{SNIPS}~\cite{38}, \textbf{AS-IPS}~\cite{saito2020asymmetric}, \textbf{IPS-V2}~\cite{li2023propensity}, \textbf{Multi-IPS}~\cite{Multi_IPW}, and \textbf{ESCM2-IPS}~\cite{wang2022escm2}; (3) DR-based: \textbf{DR-JL}~\cite{39}, \textbf{MRDR}~\cite{41}, \textbf{DR-BIAS}~\cite{Dai-etal2022}, \textbf{DR-MSE}~\cite{Dai-etal2022}, \textbf{TDR-CL}~\cite{hxli2023litdr}, \textbf{DR-V2}~\cite{li2023propensity}, \textbf{Multi-DR}~\cite{Multi_IPW}, \textbf{RD-DR}~\cite{1}, \textbf{BRD-DR}~\cite{1}, \textbf{ESCM$^{2}$-DR}~\cite{wang2022escm2}, \textbf{KBDR}~\cite{li2024kernel}, \textbf{DCE-DR}~\cite{kweon2024doubly}, and \textbf{D-DR} \cite{ha2024finegrained}.

\subsubsection{Experiment setups.}
In this paper, we select neural collaborative filtering (NCF)~\cite{he2017neural}, a widely adopted model in debiased recommendation, as the backbone model for learning the prediction model. NCF uses a neural embedding learning approach to understand the feature embedding of users and items, modeling the user-item interaction as a combinative function with these embeddings. We use Adam as the optimizer for both the imputation and prediction models. All experiments are run on Pytorch with NVIDIA GeForce RTX TITAN V as the computational resource. We tune the learning rate in $\{0.001, 0.005, 0.01, 0.05\}$, weight decay in $\{5e-6, 1e-5, 5e-5, 1e-4, 5e-4, 1e-3, 5e-3, 1e-2\}$, the embedding size of NCF in $\{4, 8, 16, 32\}$ for \textbf{Coat} dataset and $\{16, 32, 64, 128\}$ for \textbf{Yahoo} and \textbf{KuaiRec} dataset. We tune the batch size in $\{64,128,256\}$ for \textbf{Coat} and $\{2048,4096,8192\}$ for both \textbf{Yahoo! R3} and \textbf{KuaiRec}. The sample size of the Monte Carlo method is tuned between $[100,500]$ for \textbf{Coat} and $[10,60]$ for both \textbf{Yahoo! R3} and \textbf{KuaiRec}. For propensity estimation, to ensure comparison fairness, all methods do not use unbiased data. Therefore, instead of the Naive Bayes method, we use Logistic Regression (i.e., a single-layer NCF) as the propensity model. For the neural numbers and structure of NCF, we adopt the 2k -> k -> k -> 1 structure in our experiments, where k is the embedding size of NCF.

\subsection{Evaluation Metric} 
Following the previous studies \citep{li2023removing,xiao2024addressing,38}, we employ three widely used evaluation metrics—AUC, Recall@K, and NDCG@K—to assess the debiasing performance of our model. Specifically, AUC (Area Under the Curve) measures the model's ability to distinguish between positive and negative samples by calculating the area under the ROC curve.
Recall@K evaluates the proportion of user-preferred items successfully recommended within the top K recommendations, reflecting the coverage of the recommendation system.
NDCG@K (Normalized Discounted Cumulative Gain at K) combines the relevance of recommended items with their ranking positions, applying a discount and normalization to assess the quality of the top K recommendations' ranking as below
\begin{align}
D C G_u @ K=\sum_{i \in D_{\text {test }}^u} \frac{I\left(\hat{z}_{u, i} \leq K\right)}{\log \left(\hat{z}_{u, i}+1\right)}, \\
N D C G @ K=\frac{1}{|U|} \sum_{u \in U} \frac{D C G_u @ K}{I D C G_u @ K},
\end{align}
where IDCG represents the best possible DCG, and $\hat{z}_{u, i}$ denotes the ranking of the item $i$ among all items in the test set for the user $i$. Higher NDCG represents better recommendation performance and the best possible NDCG is 1. The Recall@K can be formulated as 
\begin{align}
    & \text { Recall }_{u} @ K=\frac{\sum_{i \in D_{\text {test }}^u} I\left(\hat{z}_{u, i} \leq k\right)}{\min \left(K,\left|D_{\text {test }}^u\right|\right)}, \\
    & \text { Recall@K }=\frac{1}{|U|} \sum_{u \in U} \operatorname{Recall}_u @ K.
\end{align}
% \end{equation}
We set $K=5$ for \textbf{Coat} and \textbf{Yahoo! R3} while $K=50$ for \textbf{KuaiRec} following previous studies.

% For the latter two metrics, we evaluate using $K = 5$.
% 说清楚使用的指标,AUC,, NDCG@K
\begin{figure*}[t]
\centering
\includegraphics[width=0.99\linewidth]{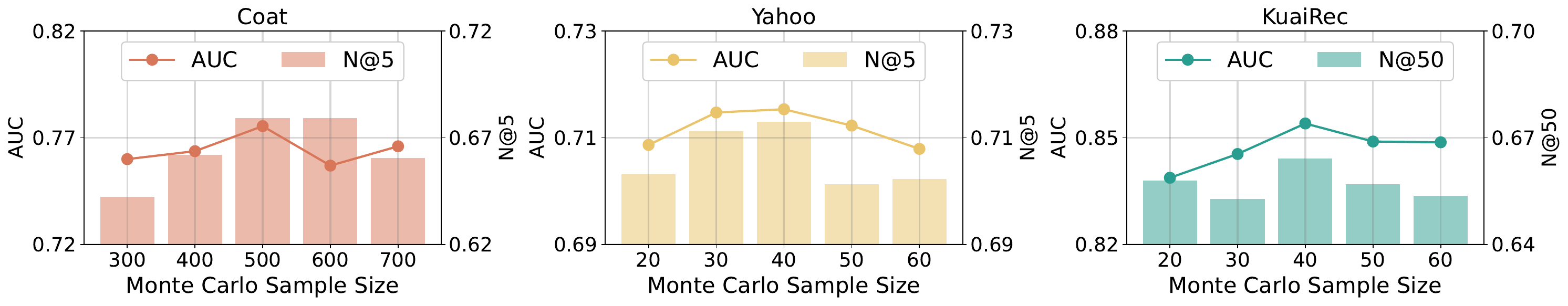}
% \vspace{-4pt}
\caption{Performance compared to baseline with varying Monte Carlo sample sizes.}
\label{fig_mc}
\end{figure*}

\begin{figure*}[t]
\centering
\includegraphics[width=0.99\linewidth]{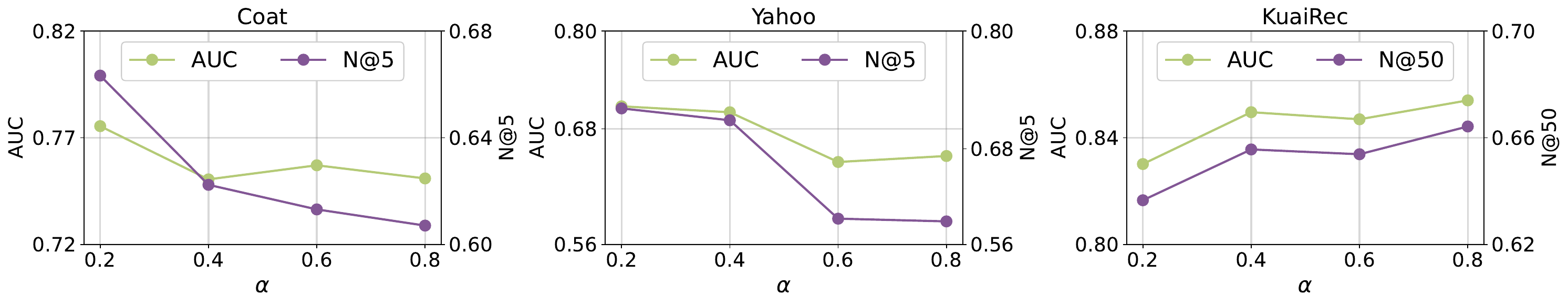}
% \vspace{-4pt}
\caption{Model performance for different value of trade-off parameter $\alpha$.}
\label{fig_alpha}
\end{figure*}

\subsection{Performance Comparison}
% To demonstrate the effectiveness of our proposed method, we give two implementations of our proposed method, one combine the naive training loss as $\mathcal{L}_{Debias}$ (-Naive) and one combine the DR loss as $\mathcal{L}_{Debias}$ (-DR). 
% We summarize the prediction performance of our proposed method with varying baselines and in \Cref{tab_main}. We can draw the following conclusions. First, most of the debiasing methods have better performance compared to naive method, which shows the necessity of debiasing. Second, our method exhibits the most competitive performance in \textbf{Coat}, \textbf{Yahoo!R3} and \textbf{KuaiRec}, significantly outperforming the baselines. This result also verifies the existence of correlated latent exogenous variables and the needs to adopt debiasing method that can handle such bias in real recommendation scenarios.
To demonstrate the effectiveness of our proposed method, we provide two implementations: one that combines the naive training loss as $\mathcal{L}_{Debias}$ (-Naive) and another that incorporates the DR loss as $\mathcal{L}_{Debias}$ (-DR). We summarize the prediction performance of our proposed method alongside various baselines in \Cref{tab_main}. From this, we can draw the following conclusions.
First, most debiasing methods perform better than the naive method, highlighting the necessity of debiasing. Second, our method shows the most competitive performance on the \textbf{Coat}, \textbf{Yahoo!R3}, and \textbf{KuaiRec} datasets, significantly outperforming the baselines. This result verifies the presence of correlated latent exogenous variables and underscores the need for debiasing methods capable of addressing such biases in real-world recommendation scenarios.

\subsection{In-Depth Analysis}
We further investigated the influence of Monte Carlo sample size on prediction performance by conducting experiments with varying sample sizes. The resulting AUC and NDCG metrics for different Monte Carlo sample sizes are depicted in \Cref{fig_mc} for all three datasets. The first observation is that a larger sample size can enhance prediction performance, which aligns with the fact that a larger sample size results in more accurate estimations when using the Monte Carlo method. Second, our method demonstrates relative stability when the sample size is sufficiently large. This suggests that our method can achieve competitive performance without requiring an excessively large Monte Carlo sample size or incurring significant computational complexity.

% \vspace{-3pt}
\subsection{Sensitivity Analysis}
We conducted a sensitivity analysis for the hyperparameter $\alpha$ described in \Cref{fn_alpha} and summarized the resulting AUC and NDCG on the \textbf{Coat}, \textbf{Yahoo!R3}, and \textbf{KuaiRec} datasets in \Cref{fig_alpha}. It is evident that on each dataset, the AUC and NDCG metrics vary significantly, indicating that the choice of $\alpha$ has a substantial impact on model performance. This may be attributed to the influence of latent exogenous variables in the data generation process. Specifically, when there is a strong correlation, the bias introduced by these latent exogenous variables may dominate, as demonstrated in \Cref{tab_mse}, which might necessitate a larger $\alpha$. Conversely, when the correlation is weak, a smaller $\alpha$ may better balance the likelihood and debiasing loss in our implementation.

% \vspace{-3pt}
\subsection{Time Complexity Comparison}
To investigate whether the Monte Carlo method significantly increases the time complexity of our proposed method, we summarize its execution time on three datasets in \Cref{tab_time}. We also provide the running time for the DR method for comparison. Our results show that our proposed method has a relatively similar execution time compared to the DR method. This may be because the summation operations involved in the Monte Carlo method can be efficiently handled by the parallel computation capabilities of modern machine learning software and platforms.
\begin{table}[t]
  \centering
  \caption{Running time (in second) comparison between DR and our proposed method on Coat, Yahoo! R3, and KuaiRec.}
    % \resizebox{0.8\linewidth}{!}{
    % \vspace{-2pt}
    \begin{tabular}{l|c|c|c}
    \toprule
    Method & \textsc{Coat}   & \textsc{Yahoo! R3} & \textsc{KuaiRec} \\ \midrule

        DR-JL & 118 & 466 & 215 \\ \midrule

        Ours & 89 & 521 & 204  \\

        \bottomrule
    \end{tabular}%
  \label{tab_time}%
  % \vspace{-2pt}
\end{table}%

\vspace{3pt}
\section{Conclusion}
\vspace{4pt}
This paper focuses on addressing selection bias in recommendation systems. Specifically, we first discuss the correlation and difference with traditional unmeasured confounding scenario. Unlike previous debiasing works that assume the independence of the exogenous variables, we model the dependence of the exogenous variables. Correspondingly, we propose a unified method that effectively handles latent exogenous variables. Our method models the data generation process, including latent exogenous variables, under mild normality assumptions and develops a Monte Carlo algorithm to numerically estimate the likelihood function based on the symmetry of the endogenous variables. Extensive experiments on synthetic datasets and three real-world datasets demonstrate the effectiveness of our method. One of the potential limitations is that the Monte Carlo sampling process, though effective, introduces computational costs that may hinder scalability for large-scale industrial systems.

%%
%% The acknowledgments section is defined using the "acks" environment
%% (and NOT an unnumbered section). This ensures the proper
%% identification of the section in the article metadata, and the
%% consistent spelling of the heading.
% \begin{acks}
% To Robert, for the bagels and explaining CMYK and color spaces.
% \end{acks}
\newpage
%%
%% The next two lines define the bibliography style to be used, and
%% the bibliography file.
\balance
\bibliographystyle{ACM-Reference-Format}
\bibliography{sample-base}

\newpage
%%
%% If your work has an appendix, this is the place to put it.
\appendix
\section{Computational details}\label{sec_ap_comp}
Firstly, we define
\begin{equation}
    z_{u,i}=g_o (x_{u,i};\theta_o )+\epsilon_i,
\end{equation}
\begin{equation}
    y_{u,i}=g_r(x_{u,i};\theta_r)+\delta_i,
\end{equation}
where $\begin{pmatrix}\epsilon_i\\\delta_i\end{pmatrix}\sim\mathcal{N}\left(\begin{bmatrix}
    0\\0
\end{bmatrix},\begin{bmatrix}1,\rho\\\rho,1\end{bmatrix}\right)$. Then, the joint probability density function of $z_{u,i}$ and $y_{u,i}$ can be expressed as
\begin{align}
    &f(z_{u,i}=z,y_{u,i}=y|x_{u,i}=x)\\
    =&\frac{1}{2\pi\sqrt{1-\rho^2}}\exp\{-\frac{1}{2(1-\rho^2)}[(z-g_o(x;\theta_o))^2\\
    &-2\rho(z-g_o(x;\theta_o)(y-g_r(x;\theta_r))+(y-g_r(x;\theta_r))^2]\},
\end{align}
Then we define the observation status as
\begin{equation}
    O_i=\begin{cases}
    1\quad if\quad z_{u,i}>0\\
    0\quad if\quad z_{u,i}\le0
    \end{cases}.
\end{equation}
Thus, the likelihood of the observational data can be written as
\begin{align}
    \mathcal{L}=&\prod_{O_i=1}f(O_i=1,y_{u,i}=y|x_{u,i}=x)\prod_{O_i=0}f(O_i=0|x_{u,i}=x)\\
    =&\prod_{O_i=1}f(z_{u,i}>0,y_{u,i}=y|x_{u,i}=x)\prod_{O_i=0}f(z_{u,i}\le0|x_{u,i}=x),
\end{align}
where
\begin{equation}
    f(z_{u,i}\le0|x_{u,i}=x)=f(\epsilon_i\le -g_o(x;\theta_o)|x_{u,i}=x)=\Phi(-g_o(x;\theta_o)),
\end{equation}
and
\begin{align}
    &f(z_{u,i}>0,y_{u,i}=y|x_{u,i}=x)\\
    =&\int_0^{\infty}\frac{1}{2\pi\sqrt{1-\rho^2}}\exp\{-\frac{1}{2(1-\rho^2)}[(z-g_o(x;\theta_o))^2\\
    &-2\rho(z-g_o(x;\theta_o)(y-g_r(x;\theta_r))+(y-g_r(x;\theta_r))^2]\}dz\\
    % \propto&\int_0^{\infty}\frac{1}{\sqrt{1-\rho^2}}\exp\{-\frac{1}{2(1-\rho^2)}[(z-g_o(x;\theta_o))^2\\
    % &-2\rho(z-g_o(x;\theta_o)(y-g_r(x;\theta_r))+(y-g_r(x;\theta_r))^2]\}dz\\
    =&\int_{-g_o(x;\theta_o)}^{\infty}\frac{1}{2\pi\sqrt{1-\rho^2}}\exp\{-\frac{1}{2(1-\rho^2)}[p^2\\
    &-2\rho p(y-g_r(x;\theta_r))+(y-g_r(x;\theta_r))^2]\}dp\\
    =&\int_{-g_o(x;\theta_o)}^{\infty}\frac{1}{2\pi\sqrt{1-\rho^2}}\exp\{-\frac{(p-\rho(y-g_r(x;\theta_r))^2)}{2(1-\rho^2)}\\
    &-\frac{(1-\rho^2)(y-g_r(x;\theta_r)^2)}{2(1-\rho^2)}\}dp\\
    =&\exp\{-\frac{(y-g_r(x;\theta_r)^2)}{2}\}\\
    &\int_{-g_o(x;\theta_o)}^{\infty}\frac{1}{2\pi\sqrt{1-\rho^2}}\exp\{-\frac{(p-\rho(y-g_r(x;\theta_r))^2)}{2(1-\rho^2)}\}dp\\
    =&\frac{1}{\sqrt{2\pi}}\exp\{-\frac{(y-g_r(x;\theta_r)^2)}{2}\}\\
    &\int_{-\frac{g_o(x;\theta_o)+\rho(y-g_r(x;\theta_r))}{\sqrt{1-\rho^2}}}^{\infty}\frac{1}{\sqrt{2\pi}}\exp\{-\frac{q^2)}{2}\}dq\\
    =&\frac{1}{\sqrt{2\pi}}\exp\{-\frac{(y-g_r(x;\theta_r)^2)}{2}\}\Phi(\frac{g_o(x;\theta_o)+\rho(y-g_r(x;\theta_r))}{\sqrt{1-\rho^2}}),
\end{align}
which means, the log-likelihood is given by
\begin{align}
    \log\mathcal{L}=&\sum_{O_i=1}-\frac{(y-g_r(x;\theta_r)^2)}{2}\\
    &+\sum_{O_i=1}\log\Phi(\frac{g_o(x;\theta_o)+\rho(y-g_r(x;\theta_r))}{\sqrt{1-\rho^2}})\\
    &+\sum_{O_i=0}\log\Phi(-g_o(x;\theta_o)).
\end{align}
Then, the parameters $\rho,\theta_o,\theta_r$ can be estimated by solving
\begin{align}
    \min_{\rho,\theta_o,\theta_r}-\log\mathcal{L}=&\min_{\rho,\theta_o,\theta_r}\sum_{O_i=1}\frac{(y-g_r(x;\theta_r)^2)}{2}\\
    &-\sum_{O_i=1}\log\Phi(\frac{g_o(x;\theta_o)+\rho(y-g_r(x;\theta_r))}{\sqrt{1-\rho^2}})\\
    &-\sum_{O_i=0}\log\Phi(-g_o(x;\theta_o)).
\end{align}
Then, if we extend this setting to the binary case, where we observe $R_i\in\{0,1\}$ s.t.
\begin{equation}
    R_i=\begin{cases}
    1\quad if\quad y_{u,i}>0\\
    0\quad if\quad y_{u,i}\le0
    \end{cases},
\end{equation}
then the probability of $f(O_i=1, R_i=1)$ can be expressed as
\begin{align}
    &f(O_i=1, R_i=1)\\
    =&f(O_i=1, y_{u,i}>0)\\
    =&\int_0^{\infty}\frac{1}{\sqrt{2\pi}}\exp\{-\frac{(y-g_r(x;\theta_r))^2}{2}\}\\
    &\Phi(\frac{g_o(x;\theta_o)+\rho(y-g_r(x;\theta_r))}{\sqrt{1-\rho^2}})dy\\
    =&\int_{-g_r(x;\theta_r)}^{\infty}\frac{1}{\sqrt{2\pi}}\exp\{-\frac{p^2}{2}\}\Phi(\frac{g_o(x;\theta_o)+\rho p}{\sqrt{1-\rho^2}})dp\\
    =&\int_{-g_r(x;\theta_r)}^{\infty}\Phi(\frac{g_o(x;\theta_o)+\rho p}{\sqrt{1-\rho^2}})\phi(p)dp\\
    =&\int_{-g_r(x;\theta_r)}^{\infty}\Phi(\frac{g_o(x;\theta_o)+\rho p}{\sqrt{1-\rho^2}})I\{p>-g_r(x;\theta_r)\}\phi(p)dp\\
    =&\int_{-\infty}^{\infty}\Phi(\frac{g_o(x;\theta_o)+\rho p}{\sqrt{1-\rho^2}})I\{p>-g_r(x;\theta_r)\}\phi(p)dp\\
    =&\mathbb{E}_{\phi(p)}\left[\Phi(\frac{g_o(x;\theta_o)+\rho p}{\sqrt{1-\rho^2}})I\{p>-g_r(x;\theta_r)\}\right]\\
    =&f(z_{u,i}>0, y_{u,i}>0)\\
    =&\mathbb{E}_{\phi(p)}\left[\Phi(\frac{g_r(x;\theta_r)+\rho p}{\sqrt{1-\rho^2}})I\{p>-g_o(x;\theta_o)\}\right].
    % \approx&\mathbb{E}_{\phi(p)}\left[\Phi(\frac{g_o(x;\theta_o)+\rho p}{\sqrt{1-\rho^2}})\frac{1}{1+\exp\{-100(p+g_r(x;\theta_r))\}}\right]
\end{align}
% \section{computation time}

% \section{Research Methods}

% \subsection{Part One}

% Lorem ipsum dolor sit amet, consectetur adipiscing elit. Morbi
% malesuada, quam in pulvinar varius, metus nunc fermentum urna, id
% sollicitudin purus odio sit amet enim. Aliquam ullamcorper eu ipsum
% vel mollis. Curabitur quis dictum nisl. Phasellus vel semper risus, et
% lacinia dolor. Integer ultricies commodo sem nec semper.

% \subsection{Part Two}

% Etiam commodo feugiat nisl pulvinar pellentesque. Etiam auctor sodales
% ligula, non varius nibh pulvinar semper. Suspendisse nec lectus non
% ipsum convallis congue hendrerit vitae sapien. Donec at laoreet
% eros. Vivamus non purus placerat, scelerisque diam eu, cursus
% ante. Etiam aliquam tortor auctor efficitur mattis.

% \section{Online Resources}

% Nam id fermentum dui. Suspendisse sagittis tortor a nulla mollis, in
% pulvinar ex pretium. Sed interdum orci quis metus euismod, et sagittis
% enim maximus. Vestibulum gravida massa ut felis suscipit
% congue. Quisque mattis elit a risus ultrices commodo venenatis eget
% dui. Etiam sagittis eleifend elementum.

% Nam interdum magna at lectus dignissim, ac dignissim lorem
% rhoncus. Maecenas eu arcu ac neque placerat aliquam. Nunc pulvinar
% massa et mattis lacinia.

\end{document}